\documentclass[12pt,a4paper]{cibb}

\makeatletter
\providecommand{\@ordinalM}[2]{#1}
\makeatother

\usepackage{subfigure,graphicx}
\usepackage{amsmath,amsfonts,latexsym,amssymb,euscript,xr}
\usepackage{booktabs}
\usepackage[nodayofweek]{datetime}
\usepackage{hyperref}
\usepackage{fmtcount}
\usepackage[english]{datenumber}
\usepackage[absolute]{textpos}
\usepackage{comment}

\usepackage[table]{xcolor}
\usepackage{color,colortbl,tabularx}

\usepackage[english]{babel}
\usepackage[protrusion=true,expansion=true]{microtype}
\usepackage{amsmath,amsfonts,amsthm}
\usepackage{pifont}

\def\black{\color{black}}
\def\blue{\color{blue}}

\definecolor{LightBlue}{rgb}{0.88,0.9,0.9}

\title{\Large $\ $\\ \bf Latent Space Analysis for Interpretable Uncertainty in Melanoma Classification}

\author{\large Ciro Listone$^{1,^*}$ and Aniello Murano$^{1}$}
\address{\footnotesize $\ $\\$^1$ University of Naples Federico II, Naples, Italy.\\
\bigskip
ORCID codes: CL 0009-0003-5532-9188; AM 0000-0003-4876-3448.
\bigskip
\newline
$^*$corresponding author: ciro.listone@unina.it
}

\abstract{\small VAE, Content-Based Image Retrieval, Melanoma Classification, Interpretable Uncertainty \normalsize
\\[17pt]
{\bf Abstract} Melanoma is a highly aggressive skin cancer, making early and accurate diagnosis critical. While deep learning excels in skin lesion classification, standard ``black-box" models struggle to explain diagnostic uncertainty, limiting clinical trust. This work introduces a hybrid framework combining a class-aware adversarial Variational Autoencoder and an XGBoost classifier, transcending simple binary classification by leveraging a generative latent space for interpretable decision support. Guided by adversarial training, the model learns the visual characteristics of skin lesions and projects them into a continuous latent space, ensuring that similar images are grouped closely together. Trained on this latent space, the XGBoost classifier achieves a robust AUC of 0.868, competing closely with state-of-the-art models. For borderline cases, the framework enables clinicians to leverage the latent topology through Content-Based Image Retrieval. This provides a dual benefit: it allows the clinician to visually compare an ambiguous lesion against biopsy-confirmed precedents and acts as an early warning sign since a borderline classification can indicate that a lesion shares features of both nevi and melanomas, potentially requiring close monitoring. Our approach translates algorithmic hesitation into transparent, evidence-based visual support, bridging the gap between predictive performance and clinical trust.
}

\begin{document}


\thispagestyle{myheadings}
\pagestyle{myheadings}

\section{Introduction}
\label{sec:SCIENTIFIC-BACKGROUND}
Melanoma is one of the most aggressive and deadly forms of skin cancer due to its high metastatic potential and rapid progression, making it a major clinical concern~\cite{yang2020trends}. It is widely acknowledged that early detection significantly improves patient survival, highlighting the clinical urgency for advanced diagnostic tools. Traditional diagnostic methods, particularly biopsies, remain the gold standard; however, they are invasive, resource-intensive, and unsuitable for widespread screening. The proliferation of dermoscopic imaging and the rise of deep learning offer fast, non-invasive, and scalable diagnostic alternatives. Yet, despite recent advances, most existing automated skin lesion analysis models operate as unexplainable ``black boxes" centered strictly on classification. In clinical settings it is essential not just to detect malignancy, but to understand and transparently communicate diagnostic uncertainty.

\paragraph{Our contribution.} Our model bridges this gap by introducing a novel hybrid framework combining a class-aware adversarial Variational Autoencoder (VAE) and an XGBoost classifier, designed not only to classify lesions but to make uncertainty interpretable. While VAEs are traditionally employed as generative models primarily aimed at image reconstruction and synthesis, we propose an alternative paradigm. Rather than focusing on the generation of new samples, our approach repurposes the generative mathematical formulation to distill clinical features. Guided by adversarial training and an auxiliary classifier, the model learns the visual characteristics of skin lesions and projects them into a continuous latent space, ensuring that morphologically similar images are grouped closely together. This structured representation moves beyond standard feature extraction, capturing the core morphological traits of the lesions in a continuous geometric space. 
To leverage this latent space for diagnosis, we train an XGBoost classifier on the learned representations using a continuous probability score as an output. For borderline cases, where the model exhibits high uncertainty, our framework enables clinicians to leverage the latent topology through a Content-Based Image Retrieval (CBIR) system. Instead of leaving the physician with an uninterpretable percentage, the system retrieves morphologically affine historical cases with biopsy-confirmed labels directly from the learned space. This provides a dual clinical benefit: first, it allows the clinician to visually compare an ambiguous lesion against known precedents to finalize the diagnosis; second, it frames the algorithmic hesitation as a meaningful early warning. 

\paragraph{Related Work.}
While Convolutional Neural Networks (CNNs) remain the established standard in medical 
image analysis and skin lesion classification~\cite{khan2019classification}, Autoencoders 
(AEs) have emerged as a compelling alternative for learning compact, meaningful data 
representations. Composed of an encoder and a decoder, AEs compress high-dimensional 
inputs into a lower-dimensional latent space, retaining only the most salient features. 
VAEs~\cite{kingma2013auto} extend this framework by 
parameterizing the latent space as a continuous probability distribution, defined by 
a mean vector $\mu$ and a standard deviation vector $\sigma$, enabling both 
generative modeling and robust feature learning. In the biomedical domain, VAEs have 
proven effective as feature extractors across diverse tasks, such as tumor stratification 
from RNA-seq data~\cite{way2018extracting}. Within dermatology specifically, they have 
been applied to anomaly detection~\cite{lu2018anomaly} and to assessing model sensitivity 
to imaging artifacts~\cite{casti2022sensitivity}. A known limitation of standard VAEs, 
however, is the tendency to produce blurry reconstructions, which can hinder the capture 
of fine morphological detail. Integrating adversarial training into VAE-GAN 
frameworks~\cite{rais2024exploring} addresses this by compelling the model to preserve 
high-frequency visual structures, yielding richer and more discriminative latent 
representations.

\paragraph{Outline.} This paper is organized as follows: Section 2 describes the experimental dataset and details the proposed framework. Section 3 evaluates the predictive performance and the semantic consistency of our system. Section 4 concludes the work and outlines future research directions.
\section{Data and Methods}
\label{sec:DATA-AND-METHODS}
This section provides a comprehensive account of the experimental setup adopted in 
this study. We first describe the dataset used to train and evaluate the proposed 
framework, including its composition and preprocessing pipeline. We then detail the 
architectural design of the model, motivating the key design choices that characterize 
the hybrid VAE-GAN approach.

\subsection{Dataset and Preprocessing}

The images analyzed in this study were sourced from the public ISIC 2019 dataset~\cite{tschandl2018ham10000, codella2018skin, combalia2019bcn20000}. Among its eight original classes, we focused on the two most clinically critical ones: nevi (NV) and melanomas (MEL), whose high morphological overlap makes binary classification a primary challenge in dermatology.
The selected subset comprised 17,297 training images (4,522 MEL, 12,875 NV) and 3,822 test images (1,327 MEL, 2,495 NV). The training data was further split into an 85\% training set and a 15\% validation set via stratified sampling to preserve the original class distribution.
All images were preprocessed by normalizing pixel intensities to $[0,1]$ and resizing them to $128 \times 128$ pixels. The implementation was carried out in Python, using Keras, TensorFlow and Scikit-learn, and executed on the Kaggle platform with GPU acceleration.

\subsection{Model Description}
We developed a hybrid framework that integrates VAEs with GANs, augmented by an auxiliary classifier attached to the latent bottleneck. This architecture enforces the learning of class-discriminative and artifact-invariant latent representations.

\paragraph{Encoder and Latent Representation.}
The encoder maps an input image $x \in \mathbb{R}^{128 \times 128 \times 3}$ through three strided convolutional layers with LeakyReLU activations ($\alpha = 0.2$), producing two $d{=}256$-dimensional vectors: the mean $\mu$ and the log-variance $\log(\sigma^2)$. Rather than encoding the input as a single fixed point in the latent space, the encoder thus defines a probability distribution $q_\phi(z|x) = \mathcal{N}(\mu, \sigma^2 I)$ over possible latent configurations, where $\phi$ denotes the encoder's trainable parameters and $I$ represents the $d \times d$ identity matrix. This probabilistic formulation is what distinguishes VAEs from 
standard autoencoders, as it imposes a structured, continuous topology on the latent space.
During training, a latent vector $z$ must be sampled from this distribution to be passed to the decoder. 
However, sampling is an inherently stochastic operation and, as such, is non-differentiable: gradients 
cannot flow through a random node, which would prevent the encoder from being updated via 
backpropagation. To resolve this, we employ the \textit{reparameterization trick}, which reformulates 
the sampling process to separate the source of stochasticity from the network's trainable parameters. 
Concretely, rather than sampling $z$ directly from $\mathcal{N}(\mu, \sigma^2)$, we first draw an 
auxiliary noise vector from a fixed standard normal distribution, $\epsilon \sim \mathcal{N}(0, I)$ 
and then compute the latent vector deterministically as:
\[
z = \mu + \sigma \odot \epsilon
\]
where $\odot$ denotes the element-wise product. Under this formulation, the randomness is 
entirely absorbed by $\epsilon$, which acts as an external constant with respect to the network's 
parameters during any given forward pass. The encoder outputs $\mu$ and $\sigma$ deterministically
and gradients can propagate back through both quantities unobstructed. Intuitively, the trick 
relocates the stochastic node outside the computational graph, while preserving the sampling 
semantics: $z$ is still a valid sample from $\mathcal{N}(\mu, \sigma^2 I)$, but it is now expressed 
as a differentiable function of the network's parameters. The resulting latent space is continuous 
and locally smooth, a property essential for the similarity-based retrieval tasks performed 
downstream.

\paragraph{Generative Decoder and Adversarial Training.}
The decoder reconstructs the input image $\hat{x} \in [0,1]^{128 \times 128 \times 3}$ from 
the sampled latent vector $z$ via a dense projection onto an $8 \times 8 \times 512$ tensor, 
followed by four consecutive blocks of bilinear upsampling, convolution, Batch Normalization, and LeakyReLU activation. A final convolutional 
layer with sigmoid activation maps the resulting feature maps to a three-channel output.
While this decoder successfully generates an image, a known limitation of standard VAEs is their tendency to produce blurry reconstructions.
To address this, an adversarial discriminator is introduced: a convolutional network comprising 
three strided layers, Batch Normalization and a sigmoid output, trained to distinguish real images from reconstructions. The decoder and 
discriminator engage in an adversarial competition, the former learning to generate 
increasingly realistic outputs, the latter to detect them. This adversarial pressure 
compels the decoder to preserve high-frequency morphological details critical for 
dermatological assessment.

\paragraph{Auxiliary Classification and Loss Function.}
To enforce a semantically structured latent space, an auxiliary Multi Layer Perceptron (MLP) is attached to 
$\mu$, consisting of a dense layer (64 units, LeakyReLU), a Dropout layer (rate $= 0.3$) 
and a sigmoid output. By supervising the encoding process with class labels, this component 
encourages the formation of well-separated MEL/NV clusters, a property later exploited by 
both the XGBoost classifier and the CBIR system.
The framework is optimized by minimizing the following multi-objective loss:
\[
\mathcal{L}_{\text{total}} = \mathcal{L}_{\text{recon}} + \lambda_{KL}\,\mathcal{L}_{KL} 
+ \lambda_{\text{clf}}\,\mathcal{L}_{\text{clf}} + \lambda_{\text{adv}}\,\mathcal{L}_{\text{adv}}
\]
where $\mathcal{L}_{\text{recon}}$ is the Mean Squared Error (MSE) between $x$ and $\hat{x}$; $\mathcal{L}_{KL}$ 
is the Kullback-Leibler divergence acting as a regularizer that prevents the latent space from collapsing into isolated 
clusters; $\mathcal{L}_{\text{clf}}$ is the Binary Crossentropy between true labels and the 
auxiliary classifier's predictions; and $\mathcal{L}_{\text{adv}}$ is the Binary Crossentropy 
between the discriminator's output on reconstructed images and a target of ones.

\paragraph{Optimization and Training Strategy.}
A dynamic augmentation pipeline was integrated into the training loop, applying random horizontal/vertical flips and rotations (within a $10\%$ range) to enforce spatial invariance in the learned representations. The model was trained end-to-end with the Adam optimizer using asymmetric learning rates: $2 \times 10^{-4}$ for generative components and $10^{-4}$ for the discriminator. To prevent posterior collapse, $\lambda_{KL}$ was linearly annealed from $0$ to $1$ over the first 10 epochs; the remaining weights were set to $\lambda_{\text{clf}} = 50$ and $\lambda_{\text{adv}} = 150$ scaling the classification and adversarial terms to prevent the reconstruction loss from dominating the training, thereby ensuring a semantically structured and perceptually sharp latent space. Overfitting was mitigated via Early Stopping on the validation classifier accuracy, with a patience of 15 epochs and best-weight restoration, typically halting training around epoch 73 (out of a maximum of 100).

\paragraph{Latent Feature Extraction and Classification.} After training the encoder is used as a deterministic feature extractor: for each image $\mu$ is extracted, omitting the noise term $\epsilon$ to guarantee stable representations. These vectors are standardized to zero mean and unit variance before classification.
Three classifiers were evaluated on the extracted features: Support Vector Machine (SVM), Random Forest (RF) and eXtreme Gradient Boosting (XGBoost). Hyperparameters were selected via exhaustive Grid Search with 3-fold cross-validation, optimizing the macro-averaged F1-score. The final model outputs continuous probability scores, enabling threshold-independent evaluation and providing the confidence signal that can be used to trigger the downstream CBIR system for borderline cases.
\section{Results}
\label{sec:RESULTS}

Following hyperparameter optimization, our latent space classifiers were evaluated on the 
hold-out test set (Table~\ref{tab:internal_comparison}). XGBoost emerged as the most robust 
model across all reported metrics, achieving the highest accuracy (81.26\%), macro-averaged 
precision (0.79), recall (0.80) and F1-score (0.80). Notably, while Random Forest attained 
a slightly higher macro precision (0.80), it did so at the cost of a severely degraded 
recall (0.74), reflecting a strong bias toward the majority NV class. SVM exhibited a more 
balanced profile, yet consistently fell below XGBoost on every metric. These results 
confirm that XGBoost is the most effective model for extracting discriminative information 
from the learned latent representations.

\begin{table}[httb!] 
\centering
\caption{\textbf{Performance of latent space classifiers}. Evaluation of SVM, Random Forest, 
and XGBoost on the test set.}
\label{tab:internal_comparison}
    \resizebox{0.8\textwidth}{!}{
    \begin{tabular}{l c c c c}
        \toprule
          \textbf{Model} & \textbf{Accuracy} & \textbf{Macro Precision} & \textbf{Macro Recall} & 
          \textbf{Macro F1} \\
        \midrule
        \rowcolor{LightBlue} SVM & 0.7786 & 0.76 & 0.76 & 0.76 \\
        Random Forest & 0.7940 & 0.80 & 0.74 & 0.75 \\
        \rowcolor{LightBlue} \textbf{XGBoost} & \textbf{0.8126} & \textbf{0.79} & 
        \textbf{0.80} & \textbf{0.80} \\
        \bottomrule
    \end{tabular}
    }
\end{table}

To contextualize these results, we benchmarked our framework against established 
end-to-end deep learning architectures (Table~\ref{tab:sota_comparison}). To ensure a 
fair comparison, all baseline models were fine-tuned for a maximum of 100 epochs under 
the same early stopping criterion applied to our framework.

\begin{table}[httb!]
\centering
    \caption{\textbf{Comparison with State-of-the-Art models}. Evaluation of our proposed 
    framework against end-to-end convolutional architectures on the hold-out test set.}
    \label{tab:sota_comparison}
    \resizebox{0.9\textwidth}{!}{
    \begin{tabular}{l c c c c c}
        \toprule
          \textbf{Model} & \textbf{AUC} & \textbf{Accuracy} & \textbf{Macro Precision} & 
          \textbf{Macro Recall} & \textbf{Macro F1} \\
        \midrule
         \rowcolor{LightBlue} EfficientNetV2 & 0.8796 & 0.8339 & 0.82 & 0.81 & 0.81 \\
         ConvNeXt-T & 0.6644 & 0.6528 & 0.33 & 0.50 & 0.39 \\
         \rowcolor{LightBlue} ResNet50 & 0.9057 & 0.8475 & 0.84 & 0.82 & 0.83 \\
         DenseNet121 & 0.9032 & 0.8477 & 0.84 & 0.82 & 0.83 \\
        \midrule
         \rowcolor{LightBlue} \textbf{VAE+XGB (Ours)} & \textbf{0.8689} & \textbf{0.8126} 
         & \textbf{0.79} & \textbf{0.80} & \textbf{0.80} \\
        \bottomrule
    \end{tabular}
    }
\end{table}

The results reveal a heterogeneous landscape among the baselines. ConvNeXt-T exhibited a 
severe collapse under class imbalance, yielding a Macro F1-score of just 0.39 and an AUC 
of 0.6644 confirming that architectural complexity alone does 
not guarantee robustness on skewed dermatological datasets. EfficientNetV2 performed more 
competitively (AUC: 0.8796, Macro F1: 0.81), while ResNet50 and DenseNet121 achieved the 
strongest overall results (AUC $\approx 0.90$, Macro F1: 0.83).
Our VAE+XGBoost framework achieved an AUC of 0.8689 and a Macro F1 of 0.80, placing it above ConvNeXt-T and closely trailing EfficientNetV2, ResNet50, and DenseNet121. This result is particularly noteworthy given the fundamental 
difference in approach: while the baselines are end-to-end convolutional networks 
optimized directly on raw pixel data, our framework operates on a compact 256-dimensional 
latent representation learned by the VAE-GAN encoder. The slight reduction in raw predictive performance relative to the strongest baselines represents a deliberate trade-off. This marginal cost is exchanged for two critical properties that purely convolutional architectures cannot provide: a semantically interpretable latent space and the capacity for visual decision support via CBIR retrieval.
Rather than returning a discrete binary label, 
the XGBoost classifier outputs a continuous probability score $p \in [0, 1]$, directly 
quantifying predictive confidence. When this score hovers near 50\%, signaling maximum 
model uncertainty, the clinician may then choose to trigger a CBIR analysis: the query 
image is projected into the latent feature space and the $k$ most morphologically similar 
historical cases are retrieved based on their Euclidean distance.

\begin{figure}[htbp]
    \centering
    \includegraphics[width=0.9\textwidth]{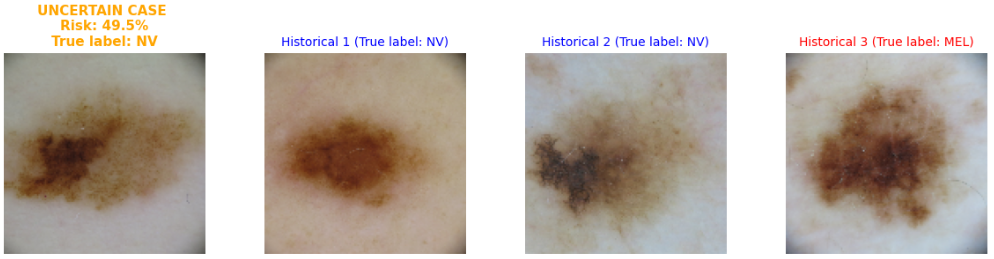}
    \caption{\textbf{Example of interpretable uncertainty via latent space retrieval.} CBIR 
    retrieval for an uncertain query lesion (left, 49.5\% risk score). The system retrieves 
    the $k=3$ nearest latent neighbors (two NV, one MEL).}
    \label{fig:uncertainty_example}
\end{figure}

A concrete example is illustrated in \autoref{fig:uncertainty_example}: for a query lesion 
assigned a risk score of 49.5\%, the system retrieves the $k=3$ most similar historical 
cases, successfully identifying two confirmed benign nevi and one confirmed melanoma sharing 
analogous visual patterns. Rather than confronting the clinician with an opaque algorithmic 
output, the system provides semantically grounded visual context: the uncertain lesion is 
presented alongside historical cases with known diagnoses and biopsy outcomes, enabling a 
more informed and evidence-based decision. 
To validate semantic consistency across the entire test set, we evaluated the CBIR system using Mean Precision@k (mP@k), which measures the average proportion of retrieved cases sharing the query's true diagnosis, and Top-$k$ Retrieval Accuracy. At $k=5$, the system achieved an overall mP@5 of 72.0\% and a Top-5 Accuracy of 93.9\% (reaching 97.1\% at $k=10$), ensuring clinicians almost always receive at least one correct biopsy-confirmed historical match.
To ensure metrics were not inflated by the benign majority, we evaluated exclusively on true melanoma queries, achieving an mP@5 of 44.7\%. Since melanomas constitute only 26.0\% of the training archive, this substantially outperforms naive random retrieval. By cutting through majority-class noise to retrieve $>2$ confirmed melanomas per top-5 search, the system provides a mathematically validated visual ``red flag'' for clinicians.
\section{Conclusion}
\label{sec:CONCLUSIONS}
In this study, we presented a VAE-GAN and XGBoost framework designed to mitigate diagnostic uncertainty in automated skin lesion analysis. Quantitatively, the system achieves highly competitive predictive performance (AUC of 0.8689, Macro F1 of 0.80), closely trailing state-of-the-art convolutional architectures, while also demonstrating robustness to class imbalance. Unlike opaque ``black box" models, our approach prioritizes interpretability by leveraging a structured latent space. The classifier outputs continuous probability scores, explicitly highlighting borderline cases where algorithmic confidence is low. To address this uncertainty, the model's architecture enables a powerful additional feature: Content-Based Image Retrieval. Clinicians can actively query the latent space to retrieve visually and semantically similar historical cases. By providing this comparative visual context, the system keeps the clinician in the loop, serving as a robust decision support tool that enhances human diagnostic confidence rather than forcing a blind binary prediction.

Future research will expand this framework into a multi-class diagnostic environment. By integrating other critical ISIC dataset classes, particularly Basal Cell Carcinomas (BCC) and Benign Keratosis-like Lesions (BKL), the system will address broader challenges in dermatological differential diagnosis. Furthermore, we aim to optimize the architecture by exploring advanced variants such as $\beta$-VAE or VQ-VAE. These improvements are expected to enhance latent feature disentanglement, yielding sharper image reconstructions and more precise nearest-neighbor retrievals for the downstream CBIR task.

\footnotesize
\bibliographystyle{unsrt}
\bibliography{bibliography_CIBB_file.bib} 
\normalsize

\end{document}